# Deploying scalable traffic prediction models for efficient management in real-world large transportation networks during hurricane evacuations

Qinhua Jiang, Brian Yueshuai He, Changju Lee, and Jiaqi Ma, *Senior Member, IEEE*

*Abstract*— Accurate traffic prediction is vital for effective traffic management during hurricane evacuation. This paper proposes a predictive modeling system that integrates Multilayer Perceptron (MLP) and Long-Short Term Memory (LSTM) models to capture both long-term congestion patterns and short-term speed patterns. Leveraging various input variables, including archived traffic data, spatial-temporal road network information, and hurricane forecast data, the framework is designed to address challenges posed by heterogeneous human behaviors, limited evacuation data, and hurricane event uncertainties. Deployed in a real-world traffic prediction system in Louisiana, the model achieved an 82% accuracy in predicting long-term congestion states over a 6-hour period during a 7-day hurricane-impacted duration. The short-term speed prediction model exhibited Mean Absolute Percentage Errors (MAPEs) ranging from 7% to 13% across evacuation horizons from 1 to 6 hours. Evaluation results underscore the model's potential to enhance traffic management during hurricane evacuations, and real-world deployment highlights its adaptability and scalability in diverse hurricane scenarios within extensive transportation networks.

## I. INTRODUCTION

OVER the past few years, several coastal regions in the United States have encountered devastating hurricanes, leading to substantial property damage and loss of lives [1], [2]. These catastrophic events have triggered a renewed focus on the enhancement of evacuation management systems. The effectiveness of evacuations crucially depends on the guidance of evacuation routes and the management of traffic flow [3]. The accuracy of traffic congestion forecasts before hurricanes and real-time traffic state predictions during hurricanes play a pivotal role in achieving these objectives. Providing reliable traffic predictions allows individuals to make well-informed decisions about evacuating, while also enabling emergency management authorities to assess the necessity of issuing evacuation orders.

During recent hurricanes (e.g., Ida 2021), for instance, massive evacuations took place in the southern Louisiana region especially in its coastal parishes[35]. Severe congestion took place on several major evacuation routes due to mandatory evacuation orders impacting hundreds of thousands of people [36]. To mitigate the impact of congestion, traffic management agencies can implement various strategies, such as staged evacuation, hard shoulder running, route guidance, and more. However, the success of these measures depends on the accuracy of traffic prediction during the evacuation.

Modeling and predicting traffic conditions during hurricane evacuations presents three significant challenges that distinguish it from regular traffic prediction. Firstly, the inherent uncertainty of hurricane events encompasses variations in intensity, landfall locations, duration, and more [37]. All these uncertainties lead to changes in the evacuation scale, evacuation direction, evacuation route, size of the evacuating population, and so on, subsequently leading to diverse traffic patterns. Secondly, the decision-making process of evacuees during such critical events is intricate[5], [46], [47], including when to start evacuating, where to evacuate to, which evacuation route to choose, and more. Unlike daily traffic, where commuters follow relatively predictable spatial-temporal patterns, mass evacuations can yield entirely different traffic patterns. Thirdly, the scarcity of high-resolution traffic observation data during historical hurricanes compounds the challenge [39]. In specific regions, powerful hurricanes can be infrequent, resulting in limited available traffic data for evacuation analysis. Take the State of Louisiana as an example, there were only 4 major hurricanes (Category 3 or higher) making landfall on its coast from 2010 to 2023. Moreover, hurricane damage may render road sensors inoperative during evacuations, further constraining the dataset's usefulness in studying spatial-temporal traffic patterns [50].

The underlying reason why hurricane traffic is predictable lies in the direct correlation between evacuation activities (such as when to leave, where to go, and which route to take) and the hurricane's characteristics (intensity, landfall location, and

---

Manuscript was submitted on 19 April 2024. This work was supported by USDOT Federal Highway Administration under grant DTFH61-12-D-00050. (Corresponding author: Jiaqi Ma)

Qinhua Jiang is with the Department of Civil and Environmental Engineering, University of California, Los Angeles, Los Angeles, USA (e-mail: qhjiang93@ucla.edu).

Brian Yueshuai He is with the Department of Civil and Environmental Engineering, University of California, Los Angeles, Los Angeles, USA (e-mail: yueshuaihe@ucla.edu).

Changju Lee is with the United Nations Economic and Social Commission for Asia and the Pacific, Bangkok, Thailand (e-mail: lee102@un.org).

Jiaqi Ma is with the Department of Civil and Environmental Engineering, University of California, Los Angeles, Los Angeles, USA (e-mail: jiaqima@ucla.edu).

timing) along with the evacuees' conditions (location, distance to landfall, time to landfall, current traffic conditions, etc.) [46]. This implies that the hurricane's properties and the spatial-temporal features of a roadway segment significantly influence the traffic conditions on a specific roadway link at a given time. For instance, a category four hurricane is more likely to cause congestion compared to a category three hurricane, as stronger hurricanes prompt a larger population to evacuate. Additionally, a roadway link in the outbound direction near the landfall location is more prone to congestion than a link farther away, as more people evacuate from cities close to the landfall location.

Regarding traffic prediction for evacuation scenarios, many previous papers utilized travel demand modeling and traffic assignment simulations to represent the travel behaviors and results traffic flows during evacuations [6], [7], [8]. These approaches follow typical traffic model principles, using either agent-based or four-step modeling methods to reflect the travel demand generated by evacuation events and simulate the traffic in the transportation network [14], [15]. However, these models require a large amount of survey data to capture people's activity preferences and develop a series of choices models to generate travel demand, which usually requires calibrating numerous parameters for each sub-model. The nature of this type of modeling makes them a more appropriate tool to analysis existing hurricanes, but unable to respond to rapid evolving of upcoming hurricanes with variable trajectory and intensities, due to the large amount of time and computing resources required to develop these models.

On the contrary, data-driven models, which rely on massive historical data instead of physical models, have drawn more attention in recent years as an important approach in studying evacuation traffic[19], [20], [22]. With the growing number of detectors installed on roadways, data-driven approaches are more often being leveraged to model traffic flows [23]. With data-driven models, the traffic patterns hidden in historical traffic data can be learned and used to predict the traffic in the future. However, due to the scarcity of hurricanes events, the data-driven hurricane evacuation traffic analysis is still not very abundant. And existing data-driven models also have various limitations, such as focusing only on one hurricane [29], can't be used to predicted hurricane with variable properties; focusing on one or a few segments of the roadway network [30], can't provide prediction for the large-scale network; and requires explicit certain types of data to run the model training, can't be transferred to other region with only limited source of historical data.

In 2015, the Federal Highway Administration (FHWA) began developing the Integrated Modeling for Road Condition Prediction (IMRCP) system, a tool that fuses real-time and archived data with results from an ensemble of forecast and probabilistic models [33], [51]. One of the system objectives is to enhance its capabilities in traffic predictions during hurricane/tropical storm season for states along the southern coast such as Louisiana, Mississippi, and Alabama. The model proposed in this study, as the core traffic prediction module in IMRCP system, aims to address three gaps in the field of hurricane evacuation traffic prediction. First, while both long-term planning and short-term response are essential in the planning and management of hurricane evacuation events [48], current emergency traffic prediction models predominantly focus on short-term, real-time predictions due to limitations in the scope of model design. Secondly, hurricane evacuations entail the movement of a significant population across extensive roadway networks, often spanning different cities [48], [49]. However, most existing evacuation traffic models concentrate on specific key corridors or selected road segments, overlooking the broader spatial-temporal correlations across the entire network and their impact on traffic patterns. Lastly, many of the prevailing approaches exhibit a high dependence on data quality and the richness of traffic data features [29], [30] , [39]. This focus on comprehensive data requirements often neglects scenarios where a complete dataset is frequently absent in many hurricane-impacted regions, such as cases where only traffic speed data is available.

To address these gaps, this study proposes a multi-scale modeling framework utilizing both Multi-Layer Perceptron (MLP) and Long Short-Term Memory (LSTM) models. The objective is to offer a full-spectrum prediction that encompasses both long-term congestion patterns and short-term speed value predictions for a large-scale transportation network. The primary contributions of this paper are summarized as follows:

- **Proposing an integrated hurricane evacuation traffic prediction framework:** This study developed an integrated hurricane evacuation traffic prediction pipeline that encompasses the entire forecast lifecycle, including data acquisition, preprocessing, model training, and deployment. This framework has demonstrated its applicability in a real-world traffic prediction system, IMRCP, and has been proven to provide support for hurricane evacuation management and planning.
- **Facilitating multi-scale and network-wide evacuation traffic predictions:** Utilizing a link-based modeling approach augmented with spatial attributes and dynamic hurricane-related features, our model effectively captures intricate spatial-temporal dependencies between hurricane dynamics and traffic patterns within a comprehensive transportation network. This capability empowers the model to make predictions regarding both long-term congestion states and short-term traffic speeds across an entire state-level transportation network in the context of hurricane evacuation.
- **Applicability to sparse datasets:** Our model introduces a specialized feature engineering and data balancing strategy tailored for hurricane evacuation scenarios, effectively addressing historical data with limited traffic attributes and a scarcity of hurricane records. This strategic approach results in robust prediction performance, facilitating the model's applicability in scenarios characterized by the absence of comprehensive traffic or hurricane data.

The rest of this paper is structured as follows: Chapter II summarizes related work and gaps. Chapter III introduces the methodology employed in this study. Chapter IV outlines the procedure for data preparation and model setup used for training the models. In Chapter V, the prediction results and analysis are presented. Finally, Chapter VI concludes the paper and offers insights into potential directions for future research.

## II. Related Work

In the realm of evacuation traffic prediction, there are two key approaches: model-based analyses[6], [7], [8], [9], [10] and data-driven methodologies[19], [20], [21], [22], [23]. Initially, the focus is on model-based studies, exploring evacuee behavior factors. However, due to laborious and time-consuming nature, there's a shift towards data-driven methods, leveraging historical traffic data for streamlined model training and wider applicability. Challenges persist in early-stage data-driven models, leading to the exploration of advanced deep neural networks for hurricane evacuation scenarios.

### A. Model-based evacuation traffic modeling

The majority of model-based traffic evacuation analyses primarily focus on studying the evacuation decisions of evacuees from a behavioral perspective [6], [7], [8]. Key aspects influencing evacuation travel behaviors include evacuation decisions [9], [10], estimated travel time [11], departure time [12], and destination choices [13]. While numerous studies have explored the impact of individual decision-making processes during evacuation [14], [15] on resulting travel demand, these approaches mainly rely on traditional methods, such as collecting survey data, which may prove inadequate for real-time hurricane evacuation scenarios. Lindell et al. [8] summarized the modeling procedures and components of large-scale evacuation processes; however, only a few models have been validated, and they are validated mainly at regional scales [40], [41], [42]. Among model-driven approaches, only a limited number of model-based methods have addressed traffic patterns during evacuation, primarily centered on analyzing highway capacity loss during the evacuation process [16], [17]. Moreover, the process of gathering and calibrating parameters for model-based approaches often proves laborious and time-consuming, leading to a growing inclination towards utilizing data-driven methods as a viable alternative.

### B. Data-driven evacuation traffic modeling

Data-driven methods enable the prediction of future traffic states by analyzing historical traffic data to discern traffic patterns. Several factors contribute to the growing popularity of data-driven approaches: firstly, model-based approaches demand tedious, labor-intensive, and time-consuming calibration efforts [18]. Secondly, the widespread deployment of diverse traffic sensors on roadway networks facilitates the use of methods reliant on extensive archived traffic data [19], [52], [53]. Thirdly, data-driven methods offer streamlined and generalized model training and updating frameworks, easing their transferability to different regions with reduced additional model development and maintenance efforts [20]. These data-driven approaches find applications in various scenarios, such as speed prediction [21], travel time prediction [22], and traffic flow prediction [23].

However, early-stage data-driven methods predominantly rely on simplistic machine learning models such as Support Vector Machine SVM [24], K-Nearest Neighbor (KNN) [25], and Artificial Neural Network (ANN) [26]. Their limitations lie in handling large-scale networks and complex traffic patterns. As a consequence, an upward trend involves employing deep neural networks for traffic prediction, including Recurrent Neural Network (RNN) [27] ,[28], [54], Convolutional Neural Network (CNN) [29], [30], Graph Convolutional Neural Network (GCN) [31], [32], and their combined forms [45]. Nevertheless, in the context of hurricane evacuation scenarios, the scarcity of specific traffic data during hurricane periods poses additional challenges in adopting the aforementioned models, which demand extensive data for robust performance.

Although the above methods has been used in general traffic prediction field, to the best of our knowledge, only a few papers have touched using deep learning models to predict traffic during hurricane evacuation [6], [29], [39]. The problem with these studies is that they focus only on one small segment from the freeway network and only use one historical hurricane for training, validation, and testing, which lack scalability and transferability. The only paper that has studied the network-level traffic prediction issue is presented by Rahman et al. [39]. This model first introduced network-wise traffic prediction using graph convolution LSTM model. But the author only used Hurricane Irma's historical traffic data for model development and testing, therefore the model's performance is unknown if implemented in other hurricanes.

Compared with prior research, this paper introduces three significant enhancements. (1) Expanded Spatial Scope: While the work by [6] examines only two specific road segments within Florida's road network, our study extends its analysis to encompass the spatial patterns across the entire state-level network. Our model accounts for spatial correlations in traffic patterns during evacuations and integrates geographic features such as coordinates, distance to landfall, link direction, and so on. This broader spatial perspective enhances traffic management planning for local agencies at the state level. (2) Extended prediction horizon: While existing studies [6], [29], [39] primarily focus on short-term traffic prediction during hurricane evacuations, none offer solutions for predicting traffic patterns over longer horizons, such as multiple days before and after landfall. Our study addresses this gap by developing two modules tailored to different stages of traffic management during hurricane evacuations. (3) Consideration of various hurricanes: Previous studies [6], [29], [39] solely analyze traffic patterns under one specific hurricanes, such as Hurricane Irma, overlooking the diverse impacts of different types of hurricanes on traffic. In contrast, our study incorporates historical traffic data from five distinct hurricanes and integrates hurricane-related features, such as hurricane category, landfall zone, time to landfall, and distance to landfall, as input variables. These enhancements enable our model to capture variations in traffic speeds across different types of hurricanes.

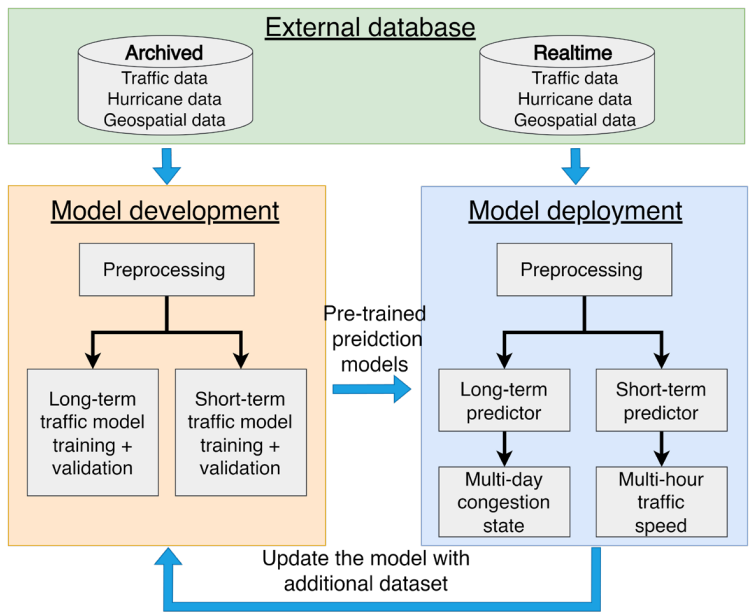

Fig. 1. The framework of the multi-scale hurricane evacuation traffic prediction model

### III. METHODS

*A. Methodology Overview*

In general traffic predictions, the prediction can be divided into short-term and long-term forecast [43]. A time span of fewer than one hour is usually considered a short-term forecast, otherwise, it is considered a medium and long-term forecast [44], [45]. In the context of emergency evacuation, however, the requirement for long-term traffic forecast is way expanded beyond one hour in the future. Due to the natural of hurricane events where hurricane is continuously changing intensity and position, the planning for hurricane evacuation events requires two-fold of supports from the traffic prediction model [48]: 1) long-term traffic information that can be generated well ahead of the hurricane (e.g., multiple days) to assist advance planning and coordination of potential support, and 2) short-term real-time prediction to provide dynamic changing in the traffic pattern (e.g., a few hours) for quick emergency response.

Based on the abovementioned two requirements, our multi-scale hurricane evacuation traffic prediction model propose two modules that tackle different aspects of the evacuation planning requirements: 1) a long-term prediction module, which provide a prediction with long span into the future with low time granularity, this helps initiate the evacuation plan days before the hurricane made landfall. and 2) a short-term model, which focuses on the near future but with high time granularity, this helps evacuation agencies to perform timely responsive operation to the most updated road traffic conditions based on the real-time traffic data fed.

Fig 1 illustrates the comprehensive framework of the hurricane evacuation traffic prediction model proposed in this paper. This model employs three primary categories of input data: traffic data, hurricane data, and network geospatial data. The input data encompasses both historical data from past hurricane evacuation events for training purposes and real-time data during active hurricane events for deployment. The model proposed in this study, being one of the core traffic prediction components within the IMRCP system [51], utilizes IMRCP's integrated data sources for both development and deployment phases. During the model development phase, multi-source historical data from previous hurricane events are extracted from the IMRCP system and leveraged to train both long-term and short-term prediction models. In the deployment phase, the model utilizes the pre-trained prediction model as a predictor and leverages real-time data from IMRCP to provide predictions for both the long-term congestion state and short-term traffic speed. After each new hurricane event, the data is archived and subsequently transferred to the model development module for the purpose of updating the prediction models.

*B. Long term congestion state prediction model*

The long-term model's primary objective is to offer multi-day predictions concerning the location and timing of congestion during a hurricane. Many existing multi-day traffic analysis studies uses 7 day, or one week as the minimum duration for traffic pattern analysis [20], [29]. Therefore, we also adopt a 7-day time span as the prediction duration for long-term congestion prediction during hurricane evacuation. To encompass traffic patterns both before and after the hurricane, the 7-day prediction horizon is defined as follows: 3 days before the landfall, the day of landfall, and 3 days after the landfall. The long-term model aims to predict the timing of congestion periods on each road segment across the hurricane-impacted 7-day range in the whole network. Therefore, we break down the 7-day time span into 28 6-hour time periods and discretize congestion state into clusters with different congestion labels, then transfer the congestion state prediction problem into a multi-class classification problem by predicting the congestion label for each of the 6-hour periods in the 7-day range.

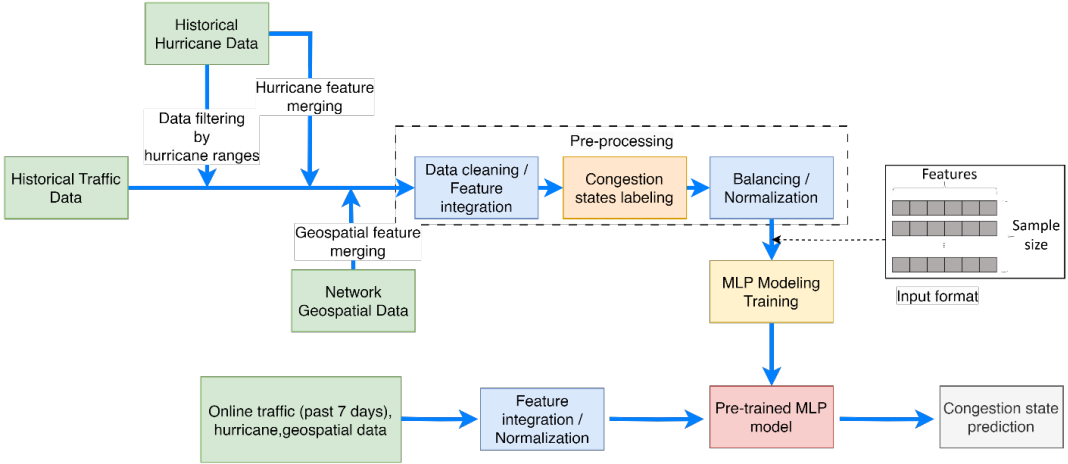

Fig. 2. Model framework for long-term hurricane evacuation traffic prediction model

Formally, the long-term congestion states prediction can be defined as a classification problem as follows:

**Input Data:** For each 6-hour period of each link segment in the road network, a set of input variables or features can be denoted as

$$X = \{X_t, X_s, X_h, X_l\} \quad (1)$$

where $X_t$ represents temporal feature set such as time of day and time to hurricane landfall, $X_s$ represents spatial feature set including link coordinates and distance to hurricane landfall, $X_h$ refers to hurricane feature set including hurricane category and landfall zone, $X_l$ represents link feature set such as link direction, number of length, and past 7-day average speed. A complete list of feature descriptions can be found in Table II of Section IV.

**Output Labels:** For each 6-hour time period of each link segment in the network, its congestion state can be denoted as a set of categorical output labels or classes:

$$Y = \{y_1, y_2, \dots y_m\} \quad (2)$$

where each $y_i$ represents a distinct congestion class or category.

The goal of the long-term prediction algorithm is to learn a function $f: X \to Y$ that can accurately map input link-time instances to their respective output congestion classes by generalizing patterns from the training data.

In this study, we adopted the MLP as the classification model for the long-term congestion state prediction. The MLP model proves particularly suitable for complex and non-linear classification tasks with intricate relationships between input features and the target class. The model's hidden layers enable it to learn intricate patterns and representations from the data, effectively handling complex decision boundaries and capturing high-dimensional feature interactions. Compared to traditional classification models like Logistic Regression, SVM, or Decision Trees, the MLP exhibits greater power in modeling complex relationships and achieving higher accuracy [6], [18], [19]. To prepare the training samples, we conduct data aggregation and annotation on original high-resolution (e.g., 5-minute interval) traffic data from historical hurricanes, identifying congestion states for each 6-hour period across the 7-day horizon. For determining the congestion state, we adopt the speed performance index (SPI) from previous traffic congestion studies [32], representing the ratio of average speed to the maximum permissible speed:

$$SPI = \frac{v_{mean}}{v_p} \times 100\% \quad (3)$$

In this paper we use the weekly average speed from the previous week to represent the maximum permissible speed. Congestion states are categorized as heavy congestion (SPI < 50%), light congestion (50% < SPI < 75%), and no congestion (SPI > 75%).

Fig 2 illustrates the model framework for the long-term congestion state prediction module. This model receives various sources of data as input, including historical traffic data, historical hurricane data, and network geospatial data. Upon integrating these diverse features, the samples are labeled using the criteria outlined in Equation 3. Subsequently, we conduct data balancing based on congestion labels to ensure the dataset's impartiality with respect to different classes. Following this, we normalize the input data and feed it into the core training module. For the multi-class classification task in this study, we employ an MLP neural network, comprising an input layer, multiple hidden layers, and an output layer.

Firstly, the input layer serves as the initial processing stage, receiving and transmitting input data to subsequent layers. Each node (neuron) in this layer corresponds to a specific feature or attribute present in the input data. The input features utilized in our model can be categorized into four distinct groups. These groups encompass link features, such as the number of lanes, directions, and non-evacuation average speed, providing essential transportation link characteristics. Spatial features, including latitude, longitude, and distance to the landfall location, contribute valuable spatial context. Temporal features, such as time of day and time to landfall, offer crucial temporal information. Lastly, the hurricane features, encompassing forecast hurricane category and potential landfall location, provide essential hurricane-related data. Secondly, the hidden layer plays a pivotal role as intermediary layers responsible for information processing. Each node in a hidden layer receives inputs from all nodes in the preceding layer and forwards its output to all nodes in the subsequent layer. This configuration allows the neural network to learn intricate patterns and relationships present in the data across different categories of features in the input layer. Lastly, the output layer generates the final predictions or outputs of the model. In our long-term prediction model, the output layer offers predictions among three congestion labels: no congestion, light congestion, and heavy congestion. This final prediction from the output layer reflects the model's evaluation of the congestion state during the hurricane evacuation process.

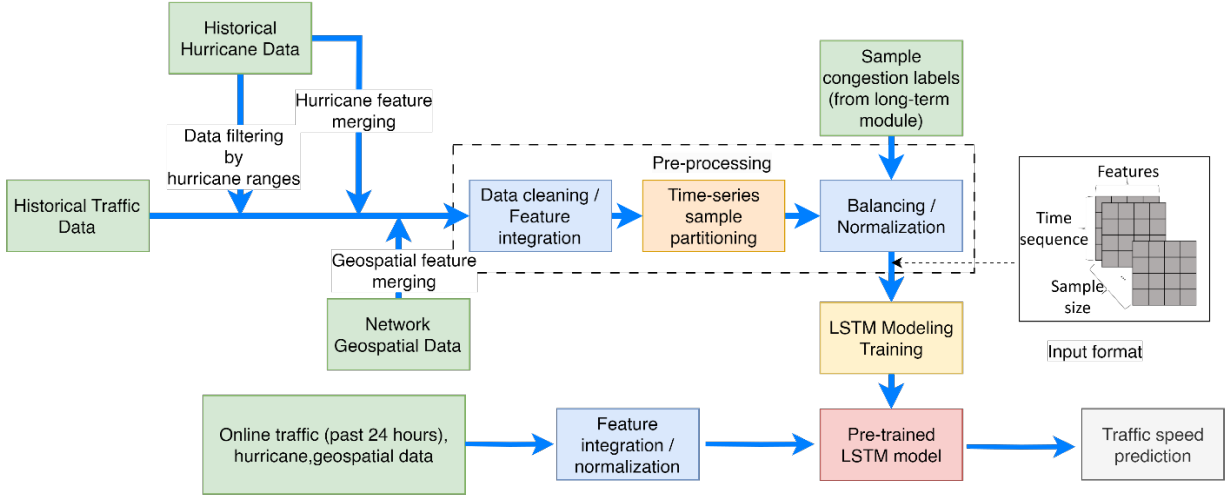

Fig. 3. Model framework for short-term hurricane evacuation traffic prediction model

*C. Short term traffic speed prediction model*

The short-term model focuses on accurately predicting speed values for each link during the 7-day hurricane impact horizon, as defined in problem statement. It aims to forecast speeds for a short-term horizon ranging from one to several hours into the future, given a particular start time. The short-term speed prediction is framed as a many-to-one time series regression problem, where the input includes time sequence data (e.g., speed data for the last 24 hours) along with link attributes, spatial-temporal attributes, and the latest hurricane attributes. The output is the predicted speed value after the specified prediction horizon from current time step.

Formally, the short-term speed prediction problem can be defined as follows:

**Input Sequences:** A sequence of historical observations of a link segment can be denoted as:

$$X = \{x_1, x_2, \ldots, x_n\} \quad (4)$$

where $x_i$ represents an individual data point at time step $i$. Each data point $x_i$ can consist of various features recorded over time 1 to $n$. These features include time-varying features such as traffic speed, time to landfall, time of day, and static features such as hurricane category, hurricane landfall zone, link coordinates, distance to landfall, and number of lanes. A complete list of feature descriptions can be found in Table II of Section IV.

**Output Prediction:** The objective is to predict a single future speed value at the next time step or next few time steps, denoted as $y$. This is a single value predicted based on the pattern or information present in the input sequence $X$.

The goal of the short-term time series speed prediction problem is to learn a mapping function $f: X \rightarrow Y$ that can capture the temporal dependencies and patterns in the historical sequence data to accurately predict the single future speed value $y$ given an input sequence $X$.

Fig 3 illustrates the model framework for the short-term traffic speed prediction module. The short-term module utilizes data resources similar to the long-term module as input. Following the integration of features, a crucial step in the short-term module involves partitioning the raw time-series data into small sequential samples of equal sequence length. To maximize the generation of samples, especially in cases of limited time-series data during historical hurricane evacuation events, we have adopted a sliding-window approach that has previously been employed in other time-series forecasting problems. This approach allows us to extract numerous small sequential data segments from the same road link sample over multiple days. Subsequently, we also perform data balancing on the partitioned samples. In this stage, we utilize pre-generated labeled data from the long-term module to ensure a balanced representation of time-series samples across different congestion states. After balancing and normalizing the pre-processed data, we then proceed to feed the data into the model training procedure.

Given the complex non-linearity and spatial-temporal dependencies between speed patterns, links, and hurricanes, the short-term prediction adopts LSTM, an advanced form of RNN that can overcome the disadvantages of RNNs such as gradient vanish when dealing with long input sequence. LSTM is specifically designed to handle sequential data, making it highly effective in capturing long-term dependencies [32]. These characteristics render LSTM a suitable choice for real-time traffic prediction during hurricane evacuation scenarios.

The cell state is a crucial component of the LSTM as it serves as a memory that runs through the entire sequence. It allows relevant information to persist over long sequences, enabling the model to capture long-term dependencies effectively. The cell state is updated at each time step using the gates and the previous cell state. In an LSTM, the cell state (hidden State) is divided into two states: short-term state ($h_t$) (similar to an RNN) and long-term state ($c_t$). The long-term state ($c_t$) stores the information to capture the long-term dependencies among current hidden state and previous hidden states over time. Traversing from the left to the right, the long-term state passes through a forget gate and drops some memories and then adds some new memories via an additional operation. A fully connected LSTM cell contains four layers (sigma and tanh) and the input vector ($x_t$) and the previous short-term state ($h_{t-1}$) are fed into these layers. The main layer uses tanh activation functions which outputs ($g_t$). The output from this layer is partially stored in long-term state ($c_t$). The other three layers are gate controller user logistic activation function and their output ranges from 0 to 1. The forget state controls which parts of the long-term state should be erased while input gate in the middle decides which parts of the input should be added. The output gate finally controls which parts of the long-term state should be read and output at this time step $y_t$.

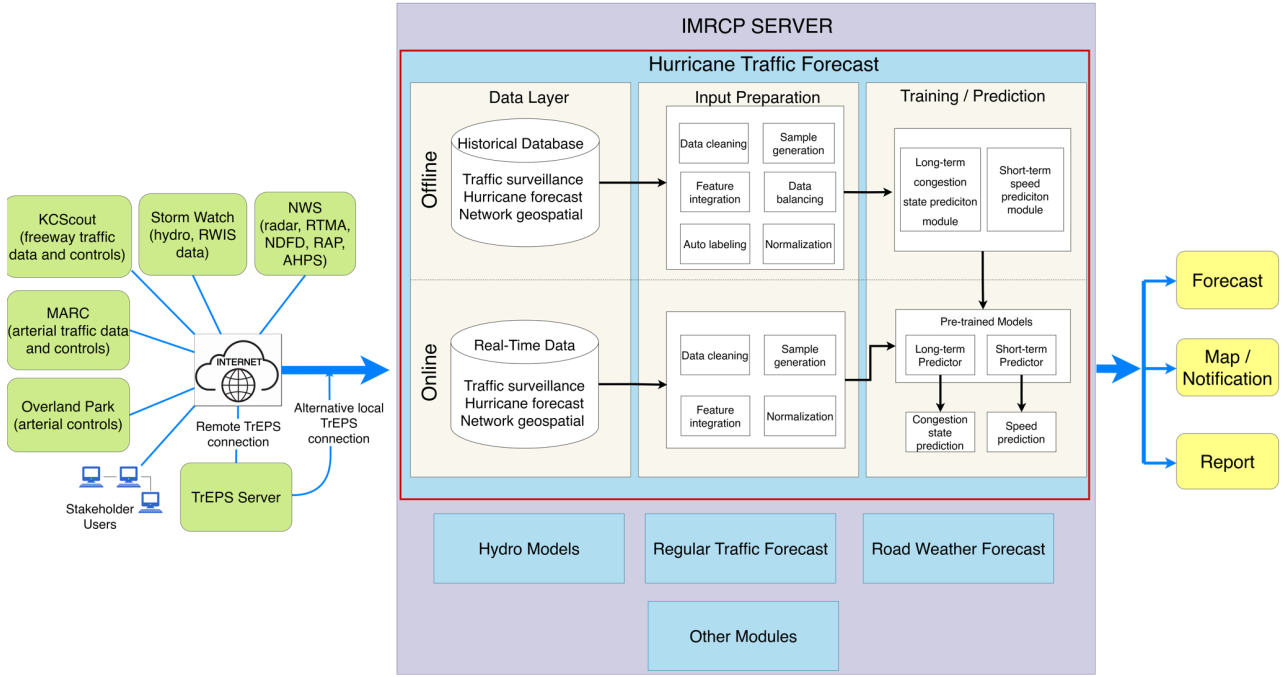

Fig. 4. IMRCP system integration diagram

Throughout the sequential processing, the LSTM iteratively updates its cell state and hidden state, considering the current input and the information from previous time steps. The final hidden state at the last time step can be used for making predictions or passed as input to other layers of the neural network for further processing. This capacity to control the flow of information through the cell state and the presence of the forget, input, and output gates enable the LSTM model to learn and capture long-term dependencies effectively, making it a powerful tool for various tasks involving sequences.

and statistic models, fusing them together in order to predict the current and future overall road/travel conditions for travelers, transportation operators, and maintenance providers. Besides different types of forecasts, IMRCP also provides flexible reporting tools and an interactive map to meet that need. The data that populates these user interface features are kept in a data store that contains collected data and data generated through forecasting components. An illustration of correlation between each component in the IMRCP system is shown in Fig 4.

The latest phase of IMRCP is engaging multiple state agencies in Louisiana, Mississippi, and Alabama to apply IMRCP capabilities in order to expand situational awareness, planning, and response to extreme weather and operational events, such as hurricane evacuations, as highlighted in the IMRCP Module in Fig 4. The Phase 5 IMRCP adopts the traffic speed prediction model proposed in this paper as the hurricane evacuation traffic speed prediction module. The model is developed based on historical traffic speed and hurricane data from the past four years in the state of Louisiana. The transportation network of Louisiana is shown in Fig 5. There are about 8,000 segmented roadway links along the evacuation routes suggested by LADOTD, which contains all interstate freeways and some US highways in Louisiana.

## IV. EXPERIMENT DESIGN

### A. Data preparation

In 2015, the FHWA Road Weather Management Program (RWMP) began developing the IMRCP system to investigate and capture the potential for operational improvements [33]. The resulting IMRCP tool incorporates real-time and archived data from various data source with results from an ensemble of forecast

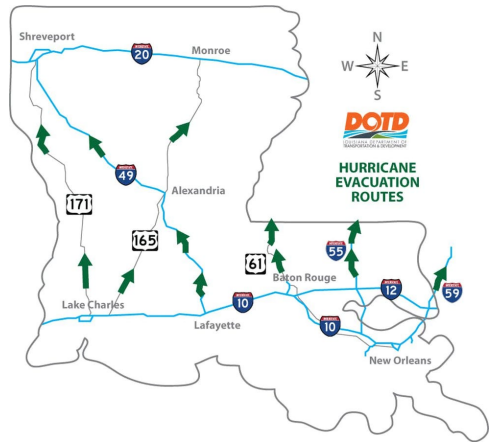

Fig. 5. Roadway network of Louisiana

The total dataset contains five separate datasets corresponding to 5 hurricanes that made landfall in Louisiana from 2019 to 2021, as shown in Table I. Each dataset covers a 7-day time range, including 3 days before landfall, the day of landfall, and 3 days after the landfall. To explore the traffic pattern during hurricane evacuations and contrast it with normal traffic conditions, we have selected two link segments of Interstate Freeway 10 (I-10) during Hurricane Ida and presented their speed data in Fig 6. Fig 6 (a) illustrates the trajectory of Hurricane Ida and the positions of the two designated links. In Figure 6 (b), we depict the historical average speed alongside the speeds recorded during the evacuation period for these two links. The historical average speed is computed by averaging the speeds observed on the corresponding days of the week over the past three months. From Fig 6 (b), two distinct patterns emerge: (1) In comparison with the historical average speed, the impact of the hurricane results in a notable congestion pattern approximately 24 to 48 hours before the landfall. The reduction in speed during evacuation surpasses the regular variation observed in the historical speed pattern. Notably, speeds during congestion can decrease to lower than 10 mph. (2) The start time and duration of congestion exhibit spatial dependencies, with the link located closer to the hurricane's path (link 2) experiencing more severe congestion. Congestion on link 2 takes place earlier than on link 1 and persists for a longer duration. Specifically, congestion on link 2 initiates around 12 PM on August 28th and persists for over 12 hours until the morning of August 29th, whereas congestion on link 1 occurs later on August 28th and is of shorter duration than on link 2. This indicates a correlation between the hurricane's impact on traffic patterns and the relative positioning of the road segment with respect to the hurricane.

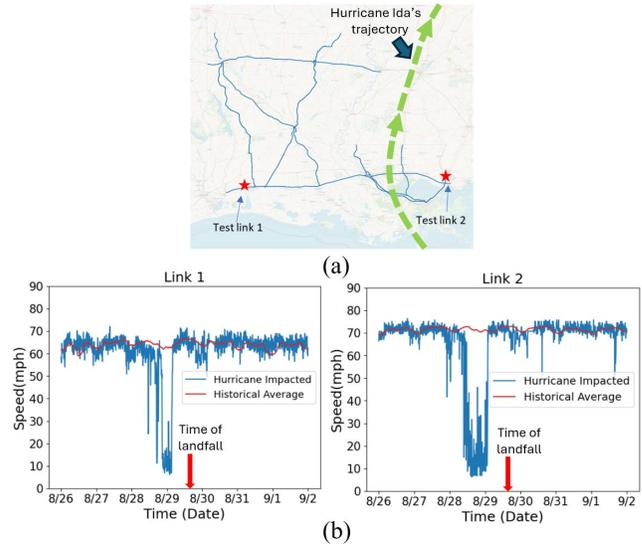

Fig. 6. Example traffic speed pattern during hurricane evacuation: (a) Link locations; (b) 7-day speed plot of example links during hurricane Ida.

To explore more general patterns of evacuation traffic, we extracted speed data from impacted road links on major evacuation routes and compared the speed differentials and congestion durations during historical hurricane evacuations. Impacted links refer to links experiencing medium to heavy congestion within a 7-day window during historical hurricanes. These impacted links were categorized into two groups based on their distance to the hurricane's path: nearby links (within 50 miles) and distant links (over 50 miles). We used 50 miles because previous studies [55], [17] have defined 50 miles as the critical distance to determine whether a location is within a hurricane's path. Table II demonstrates that both the magnitude of speed reductions and congestion durations reach their peak one day before the hurricane's landfall, gradually diminishing on the day of landfall. This indicates that hurricane-induced congestion is most pronounced in the 24 to 48 hour period preceding landfall. While congestion patterns persist until landfall, the evacuation has very limited impact on traffic in terms of speed reduction and congestion duration two days or longer before landfall. Spatially, when comparing speed differentials and congestion durations between nearby and distant links, it is evident that links closer to the hurricane's path exhibit more significant congestion patterns,

TABLE I
Dataset description

| Dataset | Hurricane Name | Hurricane Category | Landfall Time | Landfall Location | Traffic Data Source | Hurricane Data Source |
|---|---|---|---|---|---|---|
| 1 | Ida | 4 | 8/29/2021 16:55 | (29.1°N, 90.2°W) | IMRCP | NHC |
| 2 | Delta | 4 | 10/9/2020 23:00 | (29.8°N, 93.1°W) | | |
| 3 | Laura | 4 | 8/27/2020 6:00 | (29.8°N, 93.3°W) | | |
| 4 | Zeta | 3 | 10/28/2020 21:00 | (29.2°N, 90.6°W) | | |
| 5 | Barry | 1 | 7/13/2019 15:00 | (29.6°N, 92.2°W) | | |

ITSM-24-04-0117                                                                                                                                              9TABLE II
Average speed change and congestion duration of impacted links

| Day(s) before landfall | Nearby links | | Distant links | |
|---|---|---|---|---|
| | Average percentage speed change | Average duration of congestion* (hour) | Average percentage speed change | Average duration of congestion (hour) |
| 0 (landfall) | -20.9% | 3.1 | -18.7% | 2.6 |
| 1 | **-25.6%** | **6.2** | **-21.2%** | **4.7** |
| 2 | -5.2% | 0.5 | -3.7% | 0.3 |
| 3 | -0.7% | 0.1 | -0.5% | 0.1 |

characterized by greater speed reductions and longer congestion durations. The hurricane related data are extracted from National Hurricane Center (NHC) [34].

We applied feature engineering techniques to enable the model to capture crucial information from both hurricane and traffic data. To help the model better distinguish between regular speed patterns and hurricane-induced congestion patterns, we introduced two new features: the 7-day mean and standard deviation of speed. To better represent the spatial relationship between road segments and hurricane trajectories, indicating the spatial distribution of congestion during different hurricanes, we introduced a novel feature called 'distance to landfall' to denote the spatial correlation between each segment and the location where the hurricane made landfall. We also designed features such as 'time to landfall' and 'time of day' to highlight the potential relationship between the approaching hurricane and the timing of major congestions during the evacuation period, so that the model can capture the temporal characteristics of traffic patterns during hurricanes. To demonstrate the effectiveness of the feature engineering, we conducted a comparison experiment by removing two essential hurricane-related spatial-temporal features, namely time to landfall and distance to landfall. The results of this comparison can be found in Appendix B.

Table III provides the description of variables used for model training. As shown in Table III, the variables are categorized into four groups, link related, spatial, temporal, and hurricane related. The only different variables for the long-term and short-term model are the temporal variables. In the long-term model, the time-of-day variable has four classes, each represents a 6-hour period of the day, and the time to landfall variable is calculated as the number of 6-hour period from current time to the landfall time. On the other hand, for the short-term model, the temporal variables adopt hour of day instead of time of day, and the time to landfall is defined as the number of hours to the landfall time.

Utilizing the SPI defined in Equation 3, we categorized all raw data into three groups: heavy, light, and no congestion. The training dataset exhibited a high imbalance, with proportions for the three categories being 0.05, 0.12, and 0.83, respectively. To rectify this imbalance, we applied resampling to the two minority groups, specifically the heavy and light congested link samples. This resampling ensured a balanced number of samples across the various categories in the training dataset.

*B. Baseline models*

To evaluate the significance and benefits of using the MLP and LSTM we also tested a few other machine learning models as a baseline performance reference. By comparing the results of the proposed models to the baseline models, we can better understand whether the additional complexity of the new model is justified by its performance gains.

For long-term congestion pattern prediction, we selected the following models as baseline models:

- **K-nearest neighbors algorithm (KNN).** KNN is a non-parametric, supervised learning classifier, which uses proximity to make classifications or predictions about the grouping of an individual data point. While it can be used for either regression or classification problems, it is typically used as a classification algorithm, working off the assumption that similar points can be found near one another.
- **Support vector machine (SVM).** SVM is a supervised learning model used mainly for binary classification tasks, but it can be extended to multi-class classification as well. The primary objective of SVM is to find the optimal hyperplane that best separates the data points belonging to different classes in a high-dimensional feature space.

For short-term traffic speed prediction, we selected the following two models as baseline models:

- **AutoRegressive Integrated Moving Average (ARIMA)**. ARIMA is a time series forecasting model used for analyzing and forecasting time-dependent data. It is a combination of three components: AutoRegressive (AR), Integrated (I), and Moving Average (MA). ARIMA is widely used for time series forecasting when the data exhibits trends, seasonality, and autoregressive patterns.
- **Vanilla RNN model.** Vanilla RNN, also known as Simple RNN, is the basic and original form of a recurrent neural network. It is a type of artificial neural network designed for processing sequential data, such as time series, natural language, and audio data. The basic RNN has a feedback loop that allows information to persist from one time step to the next, making it suitable for sequential data.

TABLE III
Input variable description

| Variable Group | Variable Name | Description |
|---|---|---|
| Link /Traffic | Direction | Direction of the link |
| | Lanes | Number of lanes of the link |
| | Regular mean speed | The average speed on this link during the past 7 days. |
| | Regular speed standard deviation | The standard deviation of the speed on this link during the past 7 days |
| | Speed (for short-term prediction) | A sequence of traffic speed data for the past 24 hours |
| Spatial | Latitude | Latitude of the link centroid |
| | Longitude | Longitude of the link centroid |
| | Distance to landfall | Distance from the link centroid to the landfall location. (km) |
| Temporal | Time of day (long-term prediction) | It indicates which 6-hour time slot the current time falls in. 1 indicates 0:00~6:00, 2 indicates 6:00~12:00, 3 indicates 12:00~18:00, and 4 indicates 18:00~24:00 |
| | Time to landfall (long-term prediction) | Time to landfall. It indicates the time difference from landfall. In the long-term model, the time difference is calculated as the number of 6-hour time slots between the current time and landfall time. |
| | Hour of day (short-term prediction) | Hour of the day for this data record (0,1, 2, ... 22,23) |
| | Time to landfall (short-term prediction) | Time to landfall. It indicates the time difference from landfall. In the long-term model, the time difference is calculated as the number of 6-hour time slots between the current time and landfall time. |
| Hurricane | Category | Hurricane category |
| | Landfall zone | Landfall location. It's a binary variable to indicate on which half of the coastline the hurricane made landfall. 0 indicates west and 1 indicates east |
| Output | Congestion Label (long-term prediction) | Congestion status for each 6-hour period, 0 for no congestion, 1 for light congestion, 2 for heavy congestion |
| | Speed (short-term prediction) | Traffic speed 1 to 6 hours from now. |

For both the long-term and short-term prediction model, we employed a stratified 6:2:2 data split for training, validation and testing set, conducting it five times for cross-validation. In each iteration, the model trained on the current split's training data and evaluated on the test data. We average the results from these five test sets to provide a reliable measure of model performance. Table IV presents the summary of the estimated parameters for the MLP and LSTM model in our study. We implemented all the models in Python programing language. Unless otherwise specified in Table III, we have used the default parameters of PyTorch for MLP and

TABLE IV
Summary of model parameters

| Model | Parameter | Best parameters |
|---|---|---|
| MLP | Number of hidden layers | 3 |
| | Number of nodes per layer | 64 |
| | Optimizer | adam |
| | Number of epochs | 150 |
| | Batch size | 32 |
| | Learning rate | 0.001 |
| LSTM | Number of LSTM cells | 64 |
| | Number of LSTM layers | 2 |
| | Dropout | 0.5 |
| | Optimizer | adam |
| | Learning rate | 0.001 |
| | Batch size | 64 |
| | Number of epochs | 150 |

LSTM.

To select the appropriate parameters for both the long-term and short-term models, we employed the grid search method. For the MLP model, our parameter exploration included evaluating three values for the number of hidden layers (1, 2, and 3), three values for the number of nodes per layer (32, 64, 128), and three values for the batch size (32, 64, 128). On the other hand, for the LSTM model, we examined three values for the number of cells (32, 64, 128), three values for LSTM layers (1, 2, 3), and three values for the batch size (32, 64, 128). For both models, we also tested three values for learning rate (0.01, 0.001, and 0.0001). The parameter selection process explored diverse combinations to discern the impact on model performance. The finalization of parameters, as detailed in Table III, reflects the culmination of our experimentation.

*C. Evaluation Metrics*

A comprehensive set of evaluation metrics are chosen to evaluate the prediction performance of both the long-term congestion prediction model and short speed prediction model.

For the long-term speed pattern prediction model, which involves a multi-class classification problem, we utilized elements from the confusion matrix, specifically True Positives (TP), True Negatives (TN), False Positives (FP), and False Negatives (FN), to calculate the following performance metrics [62]:

- Accuracy is the most straightforward metric and represents the ratio of correctly predicted samples to the total number of samples in the dataset. It measures the overall performance of the model across all classes.

$$accuracy = \frac{TP + TN}{TP + TN + FP + FN} \quad (5)$$

- Precision, also known as Positive Predictive Value (PPV), measures the proportion of true positive predictions (correctly predicted positive samples) out of all positive predictions made by the model. It indicates how many of the positive predictions are actually correct.

$$precision = \frac{TP}{TP + FP} \quad (6)$$

- Recall is the proportion of true positive predictions (correctly predicted positive samples) out of all actual positive samples in the dataset. It indicates the model's ability to correctly identify positive samples.

$$recall = \frac{TP}{TP + FN} \quad (7)$$

- The F1 score is the harmonic mean of precision and recall. It provides a balance between precision and recall and is useful when both false positives and false negatives are equally important. The F1 score ranges from 0 to 1, with 1 being the best possible score.

$$F1\ score = \frac{2 \times precision \times recall}{precision + recall} \quad (8)$$

Note that in the evaluation of hurricane congestion prediction, recall is more important than precision. This is because ensuring the safety of evacuating residents is the top priority. Predicting congested road links allows authorities to anticipate potential bottlenecks and take proactive measures to prevent evacuees from being exposed to hazardous conditions. Therefore, our model prioritizes whether the majority of actually congested links are predicted as congested, even if this means the results come with a high number of false positives.

For the short-term model, we utilize the following metrics [63][64]:

- Root Mean Squared Error (RMSE) is a measure of the average difference between the predicted values and the actual target values. It is calculated by taking the square root of the mean of the squared differences between the predicted and actual values.

$$RMSE = \sqrt{\frac{1}{N}\sum_{i=1}^{N}(y_{true}^{i} - y_{predict}^{i})^2} \quad (9)$$

e

- Mean Absolute Error (MAE) is another measure of the average difference between the predicted values and the actual target values. It is calculated by taking the mean of the absolute differences between the predicted and actual values. Unlike RMSE, MAE does not square the errors, so it treats all errors equally regardless of their magnitude.

$$MAE = \frac{1}{N}\sum_{i=1}^{N}|y_{true}^{i} - y_{predict}^{i}| \quad (10)$$

- Mean Absolute Percentage Error (MAPE) is a relative measure of the average difference between the predicted values and the actual target values. It calculates the percentage difference between the predicted and actual values and then takes the mean of these percentage differences.

$$MAPE = \frac{1}{N}\sum_{i=1}^{N}\frac{|y_{true}^{i} - y_{predict}^{i}|}{y_{true}^{i}} \times 100 \quad (11)$$

The model proposed in this study serves as the traffic prediction module in the IMRCP Phase 4 and Phase 5 system deployed in Louisiana. To maintain conciseness, our focus in this paper is on introducing the methodology behind the model and providing a general overview of the experiments through a case study. For a more in-depth evaluation and discussion of the model deployment status and empirical results, including the integration of the traffic model with other components and a quantitative analysis of target road segments, we recommend referring to the final evaluation report [51] prepared by the IMRCP project team.

## V. RESULTS

### A. Long-term congestion state prediction

Table V presents the prediction performance of various models in forecasting long-term congestion patterns during hurricane evacuation. Our proposed MLP model stands out, outperforming all baseline models across all three congestion labels and performance metrics. This outcome highlights the efficacy of the MLP architecture, with its multiple hidden layers and activation functions, in effectively handling complex and non-linear relationships between input features and target classes, particularly in the context of hurricane congestion patterns.

TABLE V
Performance of MLP and other baseline models on long-term congestion prediction (average ± standard deviation across 5 experimental repeats).

| Model | Label | Metrics | | | |
|---|---|---|---|---|---|
| | | Precision | Recall | F1 score | Accuracy |
| KNN | No Congestion | 0.78 ± 0.011 | 0.72 ± 0.021 | 0.75 ± 0.022 | 0.75 ± 0.019 |
| | Light Congestion | 0.72 ± 0.011 | 0.76 ± 0.014 | 0.74 ± 0.021 | |
| | Heavy Congestion | 0.81 ± 0.034 | 0.77 ± 0.012 | 0.79 ± 0.014 | |
| SVM | No Congestion | 0.81 ± 0.025 | 0.80 ± 0.033 | 0.80 ± 0.039 | 0.79 ± 0.032 |
| | Light Congestion | 0.74 ± 0.009 | 0.76 ± 0.009 | 0.75 ± 0.013 | |
| | Heavy Congestion | 0.85 ± 0.012 | 0.78 ± 0.004 | 0.81 ± 0.008 | |
| MLP | No Congestion | 0.81 ± 0.021 | **0.84 ± 0.027** | 0.83 ± 0.017 | **0.82 ± 0.021** |
| | Light Congestion | 0.76 ± 0.008 | **0.83 ± 0.006** | 0.79 ± 0.018 | |
| | Heavy Congestion | 0.85 ± 0.011 | **0.86 ± 0.011** | 0.86 ± 0.025 | |

The MLP model achieves an impressive total accuracy of 82%, indicating its ability to accurately predict the locations and timings of congestion during hurricane evacuation periods. Notably, the prediction performance for heavy congestion surpasses that of light congestion in all three models (KNN, SVM, and MLP). The precision, recall, and F1 scores are consistently higher for heavy congestion compared to light congestion. This discrepancy can be attributed to the nature of heavy congestion, which is more likely to occur in specific locations, such as freeways outbound from cities, and during specific time periods, like one or two days before hurricane landfall. As a result, its spatial-temporal pattern exhibits a stronger correlation with temporal events, such as hurricane evacuations. In contrast, light congestion may occur due to various localized incidents and may not necessarily correlate directly with hurricane evacuation. Consequently, the spatial and temporal patterns for light congestion are less significant compared to heavy congestion. It is worth noting that while the MLP model only increases the overall prediction accuracy from 79% (SVM) to 82%, its impact on recall for congestion is quite significant. The recall for light congestion increases from 76% to 83%, and for heavy congestion, it improves from 78% to 86%. These enhancements indicate that despite the MLP introducing more false positives, the prediction performance for real-congested links has been greatly improved. This highlights the importance of adopting the MLP model in the traffic congestion prediction during hurricane evacuation.

To visually demonstrate the spatial distribution of hurricane-induced congestion during evacuation, we focused on a specific day (one day before hurricane Ida's landfall) to evaluate the performance of our long-term congestion pattern prediction model. Fig 7 illustrates a comparison between the ground truth and prediction results generated by the MLP model for four distinct time periods on the aforementioned day.

The left column of maps displays the actual locations of congestion throughout the four time periods, while the right column depicts the predicted congestion locations. Congestion status is represented by three colors: blue indicates no congestion, orange indicates light congestion, and red denotes heavy congestion. Analyzing the ground truth congestion maps, we observe that congestion predominantly occurs along the east-west direction of the freeway, near the Louisiana coastline. This observation suggests that the primary evacuation routes are directed either westward towards Texas or eastward towards Mississippi. Temporally, congestion initiates during the morning period (6 AM to 12 PM), intensifies during the afternoon period (12 PM to 6 PM), and persists into the evening period. This distinctive pattern differs from daily recurring traffic congestion, typically concentrated within city roadways during peak commuting hours. In contrast, hurricane-induced congestion can manifest on freeways connecting cities and endure for extended hours.

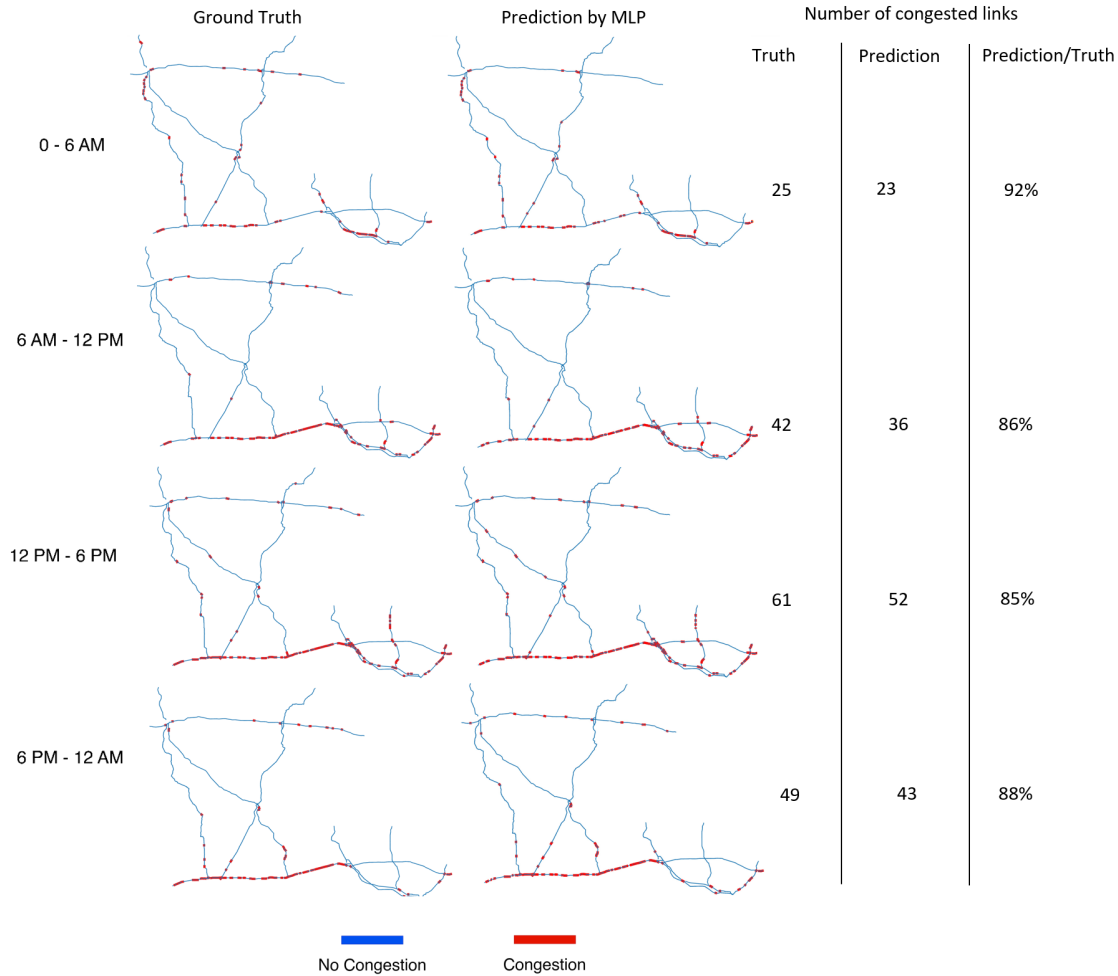

Fig. 7. Comparison of predicted and real congestion distribution one day before hurricane Ida made landfall (left column: ground truth across 4 time periods; right column: prediction across 4 time periods).

TABLE VI
Performance of LSTM and other baseline models on long-term congestion prediction (average ± standard deviation across 5 experimental repeats).

| Model | Horizon (hour) | Metrics | | |
|---|---|---|---|---|
| | | RMSE (mi/h) | MAE (mi/h) | MAPE (%) |
| ARIMA | 1 | 7.61 ± 0.31 | 6.94 ± 0.93 | 12.82 ± 1.86 |
| | 3 | 11.34 ± 0.83 | 13.93 ± 1.15 | 15.15 ± 4.07 |
| | 6 | 16.29 ± 1.27 | 17.73 ± 4.29 | 25.74 ± 6.95 |
| Vanilla RNN | 1 | 5.27 ± 0.66 | 3.75 ± 0.41 | 10.72 ± 2.54 |
| | 3 | 9.41 ± 0.27 | 10.83 ± 0.39 | 12.72 ± 3.21 |
| | 6 | 12.35 ± 1.71 | 14.61 ± 1.70 | 20.76 ± 5.03 |
| LSTM | 1 | **2.72 ± 0.06** | **2.12 ± 0.16** | **7.72 ± 2.73** |
| | 3 | **6.72 ± 0.46** | **8.76 ± 0.46** | **9.72 ± 3.09** |
| | 6 | **8.76 ± 0.80** | **8.03 ± 0.50** | **13.76 ± 4.80** |

Examining the prediction results shown in the right column maps, we find that the MLP model successfully identifies congested roadway segments during each time period, closely aligning with the actual congested locations, the predicted congested links can cover more than 85% of actual congested links across different time periods on the day before hurricane Ida made its landfall. However, it is worth noting that the predicted results occasionally show heavy congestion labels on road segments with either no congestion or only light congestion. This discrepancy suggests that the model may have overfitted the heavy congestion label compared to the other two labels, which corresponds to the recall results shown in Table V.

In conclusion, our long-term congestion pattern prediction model, based on the MLP architecture, effectively captures and predicts the spatial distribution of hurricane-induced congestion during evacuation. It demonstrates a comprehensive understanding of congestion patterns across various time periods on the day before hurricane Ida's landfall. Nonetheless, some overfitting issues were observed, especially concerning heavy congestion predictions. As a result, further refinement and optimization of the model may be required to enhance its performance and accuracy in predicting congestion levels during such critical events.

*B. Short-term traffic speed prediction*

Table VI presents a comprehensive comparison of prediction performance among different models under various prediction horizons. It is important to note that all models demonstrate high accuracy when dealing with recurring patterns and non-congested links. However, for the evaluation in Table VI, we focused exclusively on links experiencing heavy congestions throughout the 7-day hurricane period to assess their true performance in challenging scenarios.

From the results presented in Table VI, it is evident that the LSTM model consistently outperforms the other two models across all three metrics. As the prediction horizon increases, all models experience a decrease in prediction accuracy, with the errors between predicted and true speed values becoming more pronounced. Notably, the LSTM model exhibits the remarkable ability to minimize prediction errors even when dealing with a long 6-hour prediction horizon. Interestingly, when comparing the performance of LSTM with Vanilla RNN and ARIMA, it becomes apparent that the LSTM model's 6-hour prediction performance surpasses the 1-hour and 3-hour predictions of the other two models. This finding suggests that the LSTM model effectively captures sequential dependencies between previous and future timesteps, enabling it to discern correlations between hurricane features and speed patterns more adeptly than the baseline models.

The results in Table VI underscore the superiority of the LSTM model in tackling challenging congestion scenarios and extending its predictive capabilities to longer horizons. The LSTM's ability to understand and leverage sequential dependencies enables it to excel in learning the intricate relationships between hurricane features and speed patterns, setting it apart from the Vanilla RNN and ARIMA models. These findings solidify the LSTM model's potential as a valuable tool for traffic prediction tasks, especially during hurricane-induced congestion periods.

To intuitively present the performance of the short-term traffic speed prediction model, we selected two test links and plotted the continuous speed prediction over a multi-day range. As seen in Fig 6 (a), link 1 is on northbound I-59 near the border with Mississippi, and link 2 is on westbound I-10 near the border with Texas. Both links are major on major evacuation routes of Louisiana during hurricane Ida. Fig 8 illustrates the prediction performance of LSTM and baseline models on two selected test links over a three-day range, from two days before hurricane Ida made its landfall to the day of landfall. For a seven-day speed prediction comparison (three days before landfall to three days after landfall), please refer to Appendix

A. The top three plots (Fig 8 (a), (b) and (c)) represent the results for test link 1, while the bottom three plots (Fig 8 (d), (e) and (f)) correspond to test link 2. Both tests were conducted using prediction horizons of 1, 3, and 6 hours. Several key findings emerged from the analysis:

LSTM predictions exhibit reduced delay compared to ground truth when compared to baseline models. For a 1-hour prediction horizon, all three models demonstrate minimal delay in predicting congestion periods. As the prediction horizon increases, the LSTM model continues to accurately capture the start and end of congestion, while the baseline models show a substantial delay. Specifically, when the prediction horizon is extended to 6 hours, the ARIMA model can barely predict the onset of congestion, and the Vanilla RNN model's predicted onset time is 4 to 5 hours later than the actual congestion. A similar trend is observed when predicting the recovery of traffic speed from congestion. Both the ARIMA and Vanilla RNN models struggle to accurately predict the return of speed, with predicted recovery times being 2-3 hours late using a 3-hour prediction horizon and 5-6 hours late using a 6-hour prediction horizon.

The LSTM model also demonstrates superior performance in predicting the lowest speed during periods of heavy congestion. For a 1-hour prediction horizon, both LSTM and Vanilla RNN models provide reasonably accurate predictions for the lowest speed. However, as the prediction horizon increases, the performance of the baseline models significantly deteriorates, while the LSTM model maintains its accuracy with only around a 5 to 10 mi/h error on the test links. Notably, when the horizon is set to 6 hours, the speed difference between the real and predicted lowest speed is approximately 30 mi/h for the Vanilla RNN model and 45 mi/h for the ARIMA model, in contrast to the LSTM's 10 mi/h difference. These findings suggest that the baseline models are inadequate in providing valid predictions, not only in speed magnitude but also in the timing of congestion when the prediction horizon is relatively large. Additionally, the results demonstrate that the LSTM model effectively handles the spatial-temporal relationships required for short-term speed prediction, even when predicting speed values several hours into the future.

In conclusion, the LSTM model outperforms the baseline models in terms of reduced delay, accurate prediction of congestion periods, and the ability to predict lowest speeds during heavy congestion periods. The findings highlight the LSTM model's capability to handle spatial-temporal dependencies in short-term speed prediction, making it a promising approach for traffic forecasting applications.

VI. CONCLUSION

In this paper, we discussed and defined the research question of network-level traffic speed prediction during hurricane evacuation with limited traffic data scenario. A comprehensive model framework adopting the MLP and the LSTM is developed to learn the long-term congestion pattern and short-term speed pattern during hurricane evacuation. A case study using the Louisiana evacuation route network and archived speed data from 5 historical hurricanes demonstrated that the MLP long-term congestion state prediction achieved about 82% accuracy in predicting the congestion state of 6-hour period across the 7-day horizon. Additionally, the short-term speed prediction model achieved prediction MAPE from 7% to 13% for different horizons, ranging from 1 hour to 6 hours. Notably,

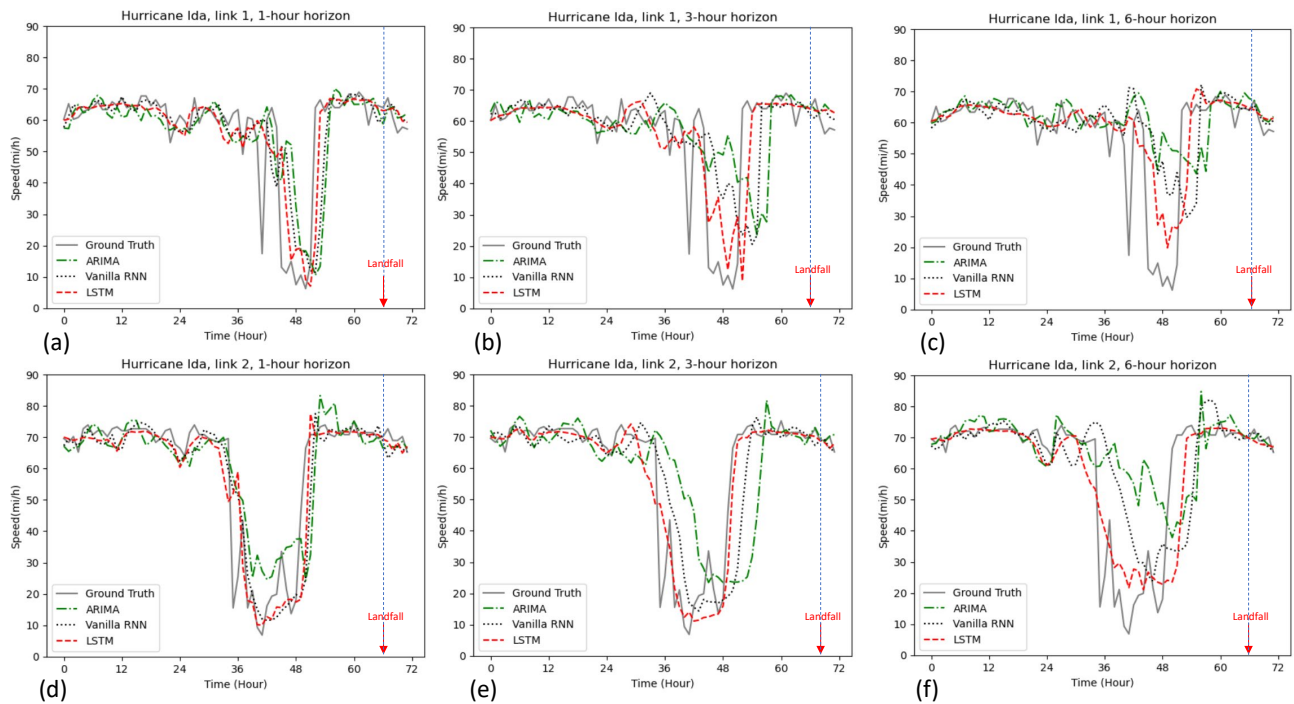

Fig. 8. Comparison of prediction performance with LSTM and baseline models during hurricane Ida ((a), (b), (c): test link 1 with 1-, 3-, and 6-hour horizon; (d), (e), (f): test link 2 with 1-, 3-, and 6-hour horizons).

ITSM-24-04-0117                                                                                                                                                                                                                                                  15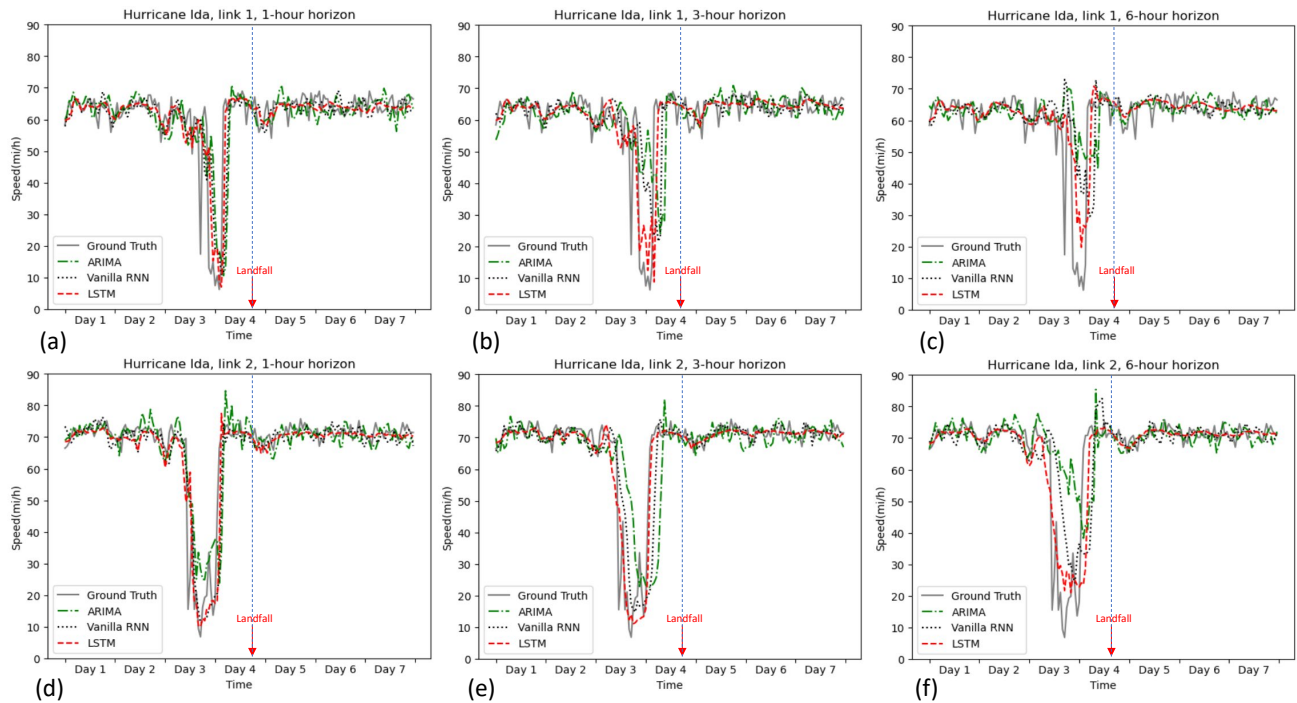

Fig. A. Comparison of prediction results between LSTM and baseline models during hurricane Ida over a 7-day range: (a), (b), (c) are results of test link 1 with 1-, 3-, and 6-hour horizon; (d), (e), (f) are results of test link 2 with 1-, 3-, and 6-hour horizons).

both the MLP and LSTM outperformed other baseline models in terms of prediction accuracy. The case study results demonstrate that the proposed model framework presents a valuable contribution for efficient traffic management by offering a holistic approach to predict traffic conditions during hurricane events in a large-scale transportation network.

As for the future research directions, firstly, it is necessary to extend the current horizon for both MLP and LSTM and explore the upper bound of the prediction performance with extender horizons. Secondly, while the current model focuses only on traffic speed due to source data limitation, we should consider incorporating the output layer with additional traffic related features if sufficient traffic data becomes available in the future.

APPENDIX A

Results of seven-day continuous speed prediction for LSTM

TABLE B-I
Performance of before and after applying feature engineering to the MLP model (average ± standard deviation across 5 experimental repeats).

| Experiment | Label | Metrics | | | |
|---|---|---|---|---|---|
| | | Precision | Recall | F1 score | Accuracy |
| Before feature engineering | No Congestion | 0.63 ± 0.092 | 0.76 ± 0.014 | 0.69 ± 0.053 | 0.66 ± 0.033 |
| | Congestion | 0.71 ± 0.038 | 0.56 ± 0.046 | 0.63 ± 0.026 | |
| After feature engineering | No Congestion | 0.81 ± 0.021 | 0.84 ± 0.027 | 0.83 ± 0.017 | 0.82 ± 0.021 |
| | Congestion | 0.81 ± 0.011 | 0.85 ± 0.011 | 0.86 ± 0.025 | |

TABLE B-II
Performance of before and after applying data balancing to the MLP model (average ± standard deviation across 5 experimental repeats).

| Experiment | Label | Metrics | | | |
|---|---|---|---|---|---|
| | | Precision | Recall | F1 score | Accuracy |
| Without balancing | No Congestion | 0.91 ± 0.041 | 0.89 ± 0.058 | 0.83 ± 0.009 | 0.85 ± 0.019 |
| | Congestion | 0.33 ± 0.032 | 0.52 ± 0.008 | 0.40 ± 0.011 | |
| With balancing | No Congestion | 0.81 ± 0.021 | 0.84 ± 0.027 | 0.83 ± 0.017 | 0.82 ± 0.021 |
| | Congestion | 0.81 ± 0.011 | 0.85 ± 0.011 | 0.86 ± 0.025 | |

TABLE C-I
Performance of MLP across 5-fold cross validation.

| Fold | Label | Metrics | | |
|---|---|---|---|---|
| | | Precision | Recall | F1 score |
| 1 | Heavy Congestion | 0.830 | 0.852 | 0.885 |
| 2 | Heavy Congestion | 0.843 | 0.862 | 0.884 |
| 3 | Heavy Congestion | 0.859 | 0.866 | 0.833 |
| 4 | Heavy Congestion | 0.847 | 0.858 | 0.838 |
| 5 | Heavy Congestion | 0.856 | 0.855 | 0.825 |
| Mean ± Std | | 0.85 ± 0.011 | 0.86 ± 0.012 | 0.86 ± 0.025 |

TABLE C-II
Performance of LSTM across 5-fold cross validation.

| Fold | Horizon (hour) | Metrics | | |
|---|---|---|---|---|
| | | RMSE (mi/h) | MAE (mi/h) | MAPE (%) |
| 1 | 6 | 8.259 | 8.132 | 14.839 |
| 2 | 6 | 8.938 | 7.573 | 15.338 |
| 3 | 6 | 8.919 | 8.768 | 13.548 |
| 4 | 6 | 8.109 | 7.664 | 13.251 |
| 5 | 6 | 8.227 | 7.795 | 13.168 |
| Mean ± Std | | 8.76 ± 0.80 | 8.03 ± 0.50 | 13.76 ± 2.80 |

TABLE D
Performance of post hurricane congestion state prediction

| Time | Label | Metrics | | | |
|---|---|---|---|---|---|
| | | Precision | Recall | F1 score | Accuracy |
| D1~D7 | No Congestion | 0.81 ± 0.021 | 0.84 ± 0.027 | 0.83 ± 0.017 | 0.82 ± 0.021 |
| | Light Congestion | 0.76 ± 0.008 | 0.83 ± 0.006 | 0.79 ± 0.018 | |
| | Heavy Congestion | 0.85 ± 0.011 | 0.86 ± 0.011 | 0.86 ± 0.025 | |
| D5~D7 (Return) | No Congestion | 0.71 ± 0.051 | 0.72 ± 0.091 | 0.71 ± 0.048 | 0.69 ± 0.049 |
| | Light Congestion | 0.68 ± 0.064 | 0.65 ± 0.067 | 0.66 ± 0.071 | |
| | Heavy Congestion | 0.67 ± 0.032 | 0.66 ± 0.056 | 0.66 ± 0.053 | |

and baseline models are presented in Fig A.

APPENDIX B

The results in Table B-I demonstrate a significant improvement in performance metrics for both no congestion and congestion labels after applying feature engineering. The total prediction accuracy increased from 0.66 to 0.82 with the incorporation of advanced features into the input layer, highlighting the positive impact of our specially designed features.

Additionally, Table B-II in Appendix B showcases the effectiveness of data balancing. Without balancing, the model's performance on congested samples significantly lags. However, after applying data balancing, the metrics for congested links are markedly improved, emphasizing the crucial role of balanced data in the training process. These comprehensive additions aim to provide a clearer understanding of our feature engineering and data balancing strategies and their positive impact on the model's performance.

APPENDIX C

A detailed breakdown of performance metrics for both the MLP and LSTM models is presented in Table C-I and Table C-II. As showcased in Table C-I and C-II, the performance remains consistent across the five folds for all metrics, encompassing both long-term and short-term models. The low variance observed in the performance metrics for both models underscores their robustness, indicating that the models are not unduly reliant on specific subsets of the data.

APPENDIX D

In analyzing this weakness, we focused on predicting the congestion states of return traffic within three days after the hurricane landfall. As depicted in Table D, the performance metrics for predicting return traffic alone exhibit a significant decline when compared to the averaged performance across the entire 7-day period. Specifically, the accuracy dropped from 80% to less than 70%, while precision, recall, and F1 score all registered a decline compared to the 7-day case.

The primary reason for this performance drop lies in the nature of return traffic, which proves to be less observable after the hurricane landfall compared to the evacuation phase. During evacuation, the population typically adheres to designated routes and specific time periods outlined in regional evacuation orders, resulting in more predictable spatial and temporal patterns. In contrast, return traffic displays greater variation and randomness, lacking a mandatory order to guide returning activities. The returning routes and timing may vary from person to person, rendering the pattern less predictable than that of evacuation traffic. This serves as a distinct weakness in our system. To address this weakness in future studies, we plan to refine the model structure and incorporate additional data sources to develop an exclusive module specifically tailored for predicting return traffic patterns.

APPENDIX E. IMRCP SYSTEM INTERFACE

As discussed in the introduction section, the system that embed the traffic model developed in this study is called the Integrated Modeling for Road Condition Prediction (IMRCP) system. The model proposed in this study functions as the traffic prediction module in the IMRCP Phase 4 and Phase 5 system deployed in Louisiana. This module collaborates with other predictive modules, such as weather prediction and road condition prediction, to support the decision-making processes of local transportation management agencies. Generally, when the traffic prediction model anticipates potential congestion on a specific link segment, local traffic management agencies

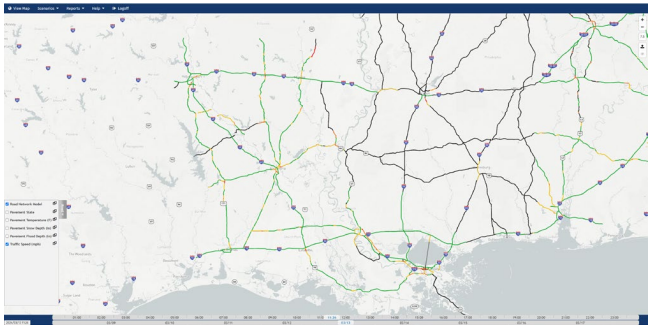
(a)

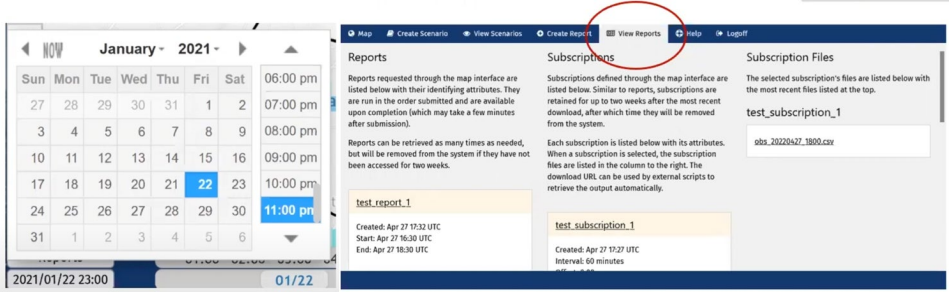
(b)  (c)

Fig E. Interface of the IMRCP system. (a) Road network and map interface (b) Variable timeframe selection (c) Report and data subscription

implement targeted strategies on the congested link or upstream links to manage traffic for enhanced safety and mobility. This may involve actions such as expanding the number of lanes through Contraflow Lane Reversal (CLR) or Hard Shoulder Running (HSR) to increase link capacity. Additionally, adjusting the speed limit by implementing Variable Speed Limit (VSL) on the congested link and upstream links notifies approaching vehicles to reduce speed in advance, improving safety. IMRCP is displayed as a web-based user interface, as shown in Fig E. It has three main components: (1) A time variable map interface which allows user to select specific historical or future timeframes to display; (2) A scenario-specific decision support tool which allows the user to define particular combination of different roadway conditions as experiment scenarios; and (3) Reports and subscription function, which allows you to create reports and data subscriptions for completed analysis.

The whole interface is a highly integrated system that provides not only traffic prediction but also road condition prediction such as surface conditions, road weather conditions and other. Specifically, during tropical storm seasons, the system will update the hurricane data from NHC and the user can select the preferred future timeframes and link segments to see the predicted traffic and road conditions through the interface.

### APPENDIX F. GENERALIZABILITY

To address the challenge that evacuation traffic pattern is expected to evolve over time and enhance the model's generalizability for unseen hurricane dataset, our approach involves adapting the design of input features based on the characteristics of new datasets. When new data becomes available, we undertake adjustments to the model's input features to best capture the emerging patterns in the dataset. The general idea is to add diverse views of data to improve model generalization and the ability to distinguish unseen classes [56][57]. For any newly introduced dataset, encompassing traffic and hurricane data from a future hurricane, we employ the following approach:

a. Handling New Categories:

If the new hurricane falls into a category that is not presented in our archived dataset, we utilize the new data to fill in the missing category in the original dataset. This step ensures a more comprehensive representation of hurricane features.

b. Enriching Feature Space:

In cases where the new data corresponds to a category already existing in our dataset, we leverage the new dataset to enrich the hurricane-related feature domain. This is achieved through two distinct approaches depending on the property of the new dataset:

1) Refining Existing Features: For instance, we extend the landfall zone from binary (west and east) to three classes (such as west, middle, and east).
2) Introducing New Features: We introduce entirely new features, such as the maximum wind speed, to further differentiate hurricanes into distinct clusters.

These adaptive approaches ensure that our model remains capable of incorporating evolving patterns in the future as new hurricane-traffic data becomes available. The continuous refinement and expansion of the model's capabilities enable it to effectively capture and adapt to the dynamic nature of evacuation traffic patterns over time.

### APPENDIX G. IMRCP SYSTEM INTERFACE

This study adopts Monte Carlo Dropout (MC Dropout), a widely adopted uncertainty estimation method, as the approach to measure the confidence interval of the prediction results. MC Dropout during inference is a technique used to estimate the uncertainty of predictions made by neural networks [58][59]. Dropout is typically used during training to prevent overfitting by randomly setting a fraction of the neurons to zero. In MC Dropout, we also applied dropout during the inference phase to generate multiple stochastic forward passes, which allows for uncertainty estimation. Here's a detailed explanation of how we implemented MC Dropout during the inference in our model:

1) During inference, keep the dropout layers active.
2) Perform N forward passes (50 in our case) through the network for the same input.
3) Collect the outputs from these passes.
4) Calculate the mean and variance of the collected outputs to get the final prediction.
5) Calculate the variance or standard deviation of the outputs to estimate the uncertainty.

For each prediction, we perform $T$ forward passes and get a set of outputs $\{\hat{y}_1, \hat{y}_2, \ldots, \hat{y}_T\}$. The mean prediction is:

$$\hat{y}_{mean} = \frac{1}{T}\sum_{t=1}^{T}\hat{y}_t$$

The standard deviation of the predictions, which provides a measure of uncertainty, is:

$$\hat{y}_{std} = \sqrt{\frac{1}{T}\sum_{t=1}^{T}(\hat{y}_t - \hat{y}_{mean})^2}$$

Assuming a normal distribution, the 95% confidence interval, $c$, for the prediction is:

$$[\hat{y}_{mean} - 1.96 \cdot \hat{y}_{std}, \hat{y}_{mean} + 1.96 \cdot \hat{y}_{std}]$$

Uncertainty measurement of prediction output is critical for a traffic speed prediction system. We introduce confidence interval calculation in predicting severe weather evacuation traffic for the following reasons [60][61]: 1) Risk assessment: computation on the range of possible traffic conditions allows authorities to assess risks more effectively and plan for worst-case scenarios, enhancing overall safety; 2) Resource allocation: Understanding which routes have higher prediction uncertainty helps in allocating resources, such as traffic personnel or signage, more effectively; 3) Dynamic decision making: Real-time updates on traffic speed predictions along with confidence intervals allow for dynamic decision-making. This adaptability is crucial when weather conditions and traffic patterns can change rapidly; and 4) Avoiding overconfidence: Without uncertainty measurements, decision-makers might rely too heavily on the predicted speeds, potentially leading to risky situations if the actual speeds are significantly different.

## ABOUT THE AUTHORS

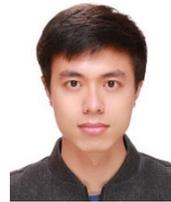

***Qinhua Jiang*** is a PhD candidate from the Civil and Environmental Engineering Department, University of California, Los Angeles. He received the B.S. and M.S. degrees in civil engineering from Beijing Jiaotong University. His areas of expertise include microscopic traffic simulation, activity-based travel demand modeling, AI/machine learning based traffic forecasting, and large-scale intelligent transportation system analysis.

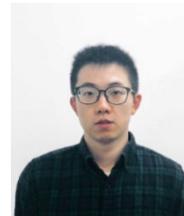

***Brian Yueshuai He*** is a Project Scientist with the Samueli School of Engineering, University of California, Los Angeles. His research interests include transportation system modeling, travel behavior analysis, and transportation planning. He has led and managed many research projects funded by federal/state/local programs in areas of multi-scale transportation system modeling, urban mobility simulation, machine learning and artificial intelligence, and computational simulation. He is a member of the TRB Sub-Committee on Travel Time, Speed, and Reliability and the International Association for Transportation Behavior Research.

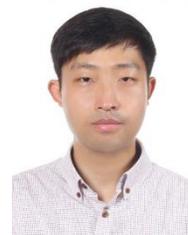

***Changju Lee*** is an economic affairs officer at the United Nations Economic and Social Commission for Asia and the Pacific. In his role at the UN, he is responsible for ITS/smart transport-related issues and their contribution towards the attainment of Sustainable Development Goals. He has a variety of experience in ITS/smart transport, transport planning, policy and economics, and data analysis. Prior to joining the UN, he was a research associate with Virginia Transportation Research Council. He holds Master degrees in City Planning and in Civil Engineering, and a Ph.D. in Civil Engineering (Transportation) from the University of Virginia.

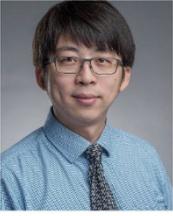
**Jiaqi Ma** (Senior Member, IEEE) is an Associate Professor with the Samueli School of Engineering, University of California, Los Angeles, and Director of FHWA Center of Excellence on New Mobility and Automated Vehicles He has led and managed many research projects funded by U.S. DOT, NSF, state DOTs, and other federal/state/local programs covering areas of smart transportation systems, such as vehicle-highway automation, intelligent transportation systems, connected vehicles, shared mobility, and large-scale smart system modeling and simulation, and artificial intelligence and advanced computing applications in transportation. He is Editor in Chief of the IEEE OPEN JOURNAL OF INTELLIGENT TRANSPORTATION SYSTEMS. He is a member of the TRB Standing Committee on Vehicle-Highway Automation, the TRB Standing Committee on Artificial Intelligence and Advanced Computing Applications, and the American Society of Civil Engineers Connected and Autonomous Vehicles Impacts Committee, and the Co-Chair of the IEEE ITS Society Technical Committee on Smart Mobility and Transportation 5.0.